\documentclass[runningheads]{llncs}
\usepackage{graphicx}
\usepackage{amsmath,amssymb} 
\usepackage{color}

\usepackage{todonotes}
\usepackage{gensymb}

\begin{document}
\pagestyle{headings}
\mainmatter

\title{Interest point detectors stability evaluation on ApolloScape dataset} 

\titlerunning{Interest point detectors stability evaluation on ApolloScape dataset}

\authorrunning{J, Komorowski et al}

\author{Jacek Komorowski \inst{1}\orcidID{0000-0001-6906-4318},
Konrad Czarnota \inst{1},
Tomasz Trzcinski \inst{1, 2}\orcidID{0000-0002-1486-8906},
Lukasz Dabala \inst{1}, 
Simon Lynen \inst{3}}
\institute{
Warsaw University of Technology, Warsaw, Poland\\
\and
Tooploox
\and 
Google
\\
}

\maketitle

\begin{abstract}
In the recent years, a number of novel, deep-learning based, interest point detectors, such as 
LIFT,
DELF,
Superpoint
or LF-Net was proposed.
However there's a lack of a standard benchmark to evaluate suitability of these novel keypoint detectors for real-live applications such as autonomous driving.
Traditional benchmarks (e.g. Oxford VGG) are rather limited, as they consist of relatively few images of mostly planar scenes taken in favourable conditions.
In this paper we verify if the recent, deep-learning based interest point detectors have the advantage over the traditional, hand-crafted keypoint detectors.
To this end, we evaluate stability of a number of hand crafted and recent, learning-based interest point detectors on the street-level view ApolloScape dataset.
\end{abstract}

\section{Introduction}

Detection of local interest points in images is the fundamental part of many computer vision applications, including 3D reconstruction~\cite{Agarwal2011}, panorama stitching~\cite{Brown2007} and monocular Simultaneous Localization and Mapping~\cite{Lynen15}. 
This topic has gained significant attention from the research community~\cite{yi2016lift,noh2017largescale,detone2017superpoint,harris1988combined,lindeberg1998feature,lowe1999object,rosten2006machine,rublee2011orb,savinov2017quad,ono2018lf}.
While traditional interest point detectors rely on hand-crafted features ~\cite{harris1988combined,lindeberg1998feature,lowe1999object},
more recent methods use machine learning techniques such as decision trees~\cite{rosten2006machine} or deep learning~\cite{yi2016lift,noh2017largescale,detone2017superpoint,savinov2017quad,ono2018lf} to train a high-performing keypoint detector. 

In the recent years, a number of novel, deep-learning based interest point detectors, such as 
LIFT~\cite{yi2016lift},
DELF~\cite{noh2017largescale},
Superpoint~\cite{detone2017superpoint} or LF-Net~\cite{ono2018lf}, was proposed.
However there's a lack of a standard benchmark, to evaluate suitability of these novel keypoint detectors for  real-live applications such as autonomous driving.
Oxford VGG~\cite{mikolajczyk2004scale}, a very popular benchmark for evaluation of local features, is rather limited, as it consists of only 8 sequences, each containing 6 images.
Images are taken in a favourable environmental conditions and contain mostly planar scenes.
The dataset does not capture variety of factors that impact the image of the visible scene, such as diverse weather conditions, different seasons or time of the day.

In the seminal work~\cite{tuytelaars2008local} on local feature detectors, the following properties of an ideal local interest points are listed: 
repeatability, distinctiveness, locality, quantity, accuracy and efficiency.
Repeatability is considered as one of the most important properties of good features.
High repeatability ensures that given two images of the same scene, taken under different viewing conditions, a high percentage of features detected on the scene part visible in both views can be found in both images.
In this paper we evaluate repeatability of a wide range of hand-crafted and learning-based interest point detectors on 
the 
ApolloScape~\cite{huang2018apolloscape} street-level view
dataset.
Due to their size and diversity this datasets allows evaluation of interest point detectors in the real-life, variable and challenging conditions.

\section{Related work}
\label{sec:related_work}

In this section we briefly review recently proposed, learning based interest point detectors.

TaSK (Task Specific Keypoint)~\cite{strecha2009training} is one of the early attempts to use machine learning methods in order to improve interest point detector performance.
It aims at improving repeatability of keypoints by learning a classifier to filter out detected keypoints to retain more stable features.
However, the method is reliant on some other interest point detector, such as DoG~\cite{lowe1999object}, and only filters out points found by the base detector.

TILDE (Temporarily Invariant Learned DEtector) \cite{verdie2015tilde} is the learning-based method to detect repeatable keypoints under drastic changes of environmental conditions such as weather and lighting.
It relies on other keypoint detector, DoG~\cite{lowe1999object}, only to generate training examples.
Training set consists of multiple stacks of images showing the same scene and taken from the same viewpoint but at different season and time of the day.
DoG features detected on training images are considered positive examples, if they are repeatable on majority images in the stack (set of images showing the same scene).
The regressor is trained to compute a value (score map) for each patch of a given size of the input image. The loss function is constructed so the learned regressor produces a peaked shape on positive samples (patches centered near good keypoint location) and small value on negative samples. 

LIFT (Learned Invariant Feature Transform) \cite{yi2016lift} is the first end-to-end, deep learning-based, method including feature detection, orientation estimation and robust descriptor extraction. The learning is based on SfM-verified DoG~\cite{lowe1999object} keypoints detected in a large collection of images.
The feature extraction pipeline is trained by presenting quadruples of image patches (base patch, positive example, negative example and a patch with no distinctive feature points)
to the network having a Siamese-like architecture with four branches.

Convolutional feature detector proposed in \cite{altwaijry2016learning} identifies regions of the input image that constitute good keypoints. 
The loss function consists of two terms: binary classification loss to classify patches as centered on the keypoint location and squared difference loss to penalize network responses on non-centered patches.
Training set is constructed by detecting interest points using a handcrafted feature detector in a sequence of images.
Tuples of points that survive SfM verification form positive training examples.



Quad-networks~\cite{savinov2017quad} is the first unsupervised interest point detection method, not relying on a hand crafted feature detector for the training set generation.
Neural network is trained to map an image patch point to a single real-valued response and then to rank points according to the response. Image patch is mapped by the neural network to a 'heat map' with ranking of each pixel.
The ranking is optimized to be repeatable under the desired transformation  classes, such as rotation, translation or intensity change.
If one point has the higher ranking than another one, it should still be higher after a transformation.
Top and bottom quantiles of the response are repeatable and can be used as an interest points.
Two training approaches are described. 
The first one uses an image dataset with ground truth data on camera poses and 3D scene structure from the LIDAR scan. Using the ground truth data, correspondence between image patches in two different images can be established.
The other approach is fully unsupervised.
The training set is constructed, by applying transformations, such as rotation and translation, to a set of base image patches.


Superpoint~\cite{detone2017superpoint}
is a self-supervised framework for training interest point detectors and descriptors. 
It operates on the full-sized images and jointly computes interest points locations and associated descriptors.
VGG-style~\cite{simonyan2014very} convolutional network is used to reduce the dimensionality of the input image. Then the network branches into interest point decoder and descriptor decoder units. 
The interest point detector is pre-trained on the synthetic data consisting of a large set of computer generated images with pseudo-ground truth interest point locations.
To boost the performance on the real data, the network is fine tuned using real images. Images are warped multiple times to help an interest point detector to see the scene from many different viewpoints and scales.

LF-Net (Local Feature Network)~\cite{ono2018lf} is a novel deep architecture for sparse matching, which can be trained end-to-end and does not require using a hand-crafted feature detector to initialize the training process.
The training requires image pairs with known relative pose and  depth maps, so the correspondence between image patches in two images can be established.
LF-Net runs the detector on the first image and finds the maxima on the produced score map. Then it optimizes the weights so that when the detector is run on the second image it produces a \emph{clean response map} with sharp maxima at right locations. 

\section{Interest point detectors evaluation}
\label{sec:interest_point_evaluation}

\subsection{Dataset}

We performed evaluation of interest point detectors on 
the recently released
ApolloScape~\cite{huang2018apolloscape} street-level view dataset.
The dataset contains almost 150 thousand frames with high quality ground truth data including pixel-level semantic segmentation, pose information and depth maps for the static background.
Image frames in the dataset are collected every one meter by the acquisition system with resolution 3384 x 2710. 
The dataset contains images taken in diverse locations under the varying weather conditions and with challenging lightning conditions (e.g. dark or very bright).

The data is acquired using Riegl VMX-1HA acquisition system consisting of two VUX-1HA laser scanners (360\degree FOV, range from 1.2m up to 420m), VMX-CS6 camera system (two front cameras with resolution 3384 x 2710),  and the measuring head with IMU/GNSS (position accuracy 20 $\sim$ 50mm, roll and pitch accuracy 0.005\degree , and heading accuracy 0.015\degree). 

\begin{figure}[t]
\centering
\includegraphics[width=0.5\textwidth]{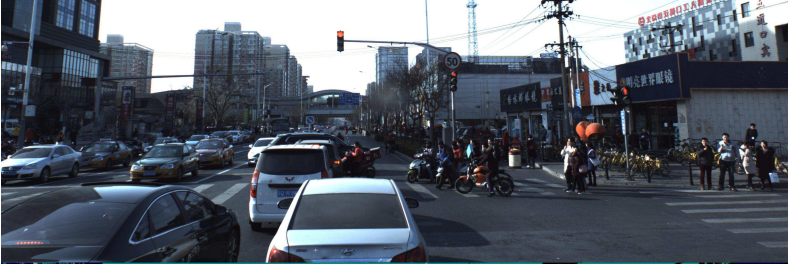}
\includegraphics[width=0.5\textwidth]{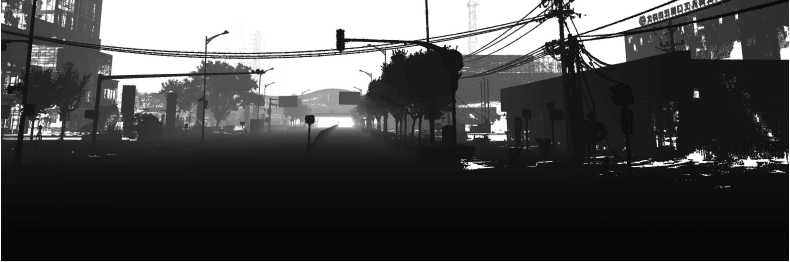}
\caption{Exemplary image (top) and corresponding depth map for the static background (bottom) from the ApolloScape~\cite{huang2018apolloscape} dataset.}
\label{fig:apolloscape}
\end{figure}

\subsection{Evaluation criteria}

We evaluate stability of the interest point detectors using \emph{repeatability} criteria introduced in~\cite{schmid1998comparing}.
\emph{Repeatability} score for an interest point detector operating on a pair of images is defined as a ratio between the number of point-to-point correspondences that can be established between detected points and the number of points detected in both images.

Computing repeatability, in general case, is not a trivial task.
First, correspondence between interest points on two images must be established.
In standard benchmarks, such as Oxford VGG Affine Covariant Regions~\cite{mikolajczyk2004scale} or HPatches~\cite{balntas2017hpatches}, this relation is defined by a homography, as they contain images of planar scenes or camera motion is purely rotational.
In general case, where the visible scene is not planar and the camera movement is not restricted, establishing such correspondence is not easy.
Second, an interest point detector in general will not detected an interest point at the exact position of the corresponding point, but at some close neighbourhood.
Third, some interest points cannot be repeated as some  parts of the scene can be observed by only one camera.

When interest point correspondence is related by the homography $H$, repeatability definition introduced in~\cite{schmid1998comparing} is commonly used.
Let $P$ be a 3D point and $p_1 = M_1 P$, $p_2 = M_2 P$ its projections of $P$ onto the image $I_1$ and $I_2$ respectively.
$M_1$ and $M_2$ are projections matrices related with images $I_1$ and $I_2$.
Two points $p1 \in I_1$ and $p2 \in I_2$ correspond, if they are related by homography $H$, that is if $dist(H p_1, p_2) < \theta$, where 
$\theta$ is some fixed threshold.
$d_1$ and $d_2$ denote points that could be potentially detected on both images, defined as as $d_1 = \left\{ p_i \in I_1 | H_{21}p_i \in I_2 \right\}$ and
$d_2 = \left\{ p_2 \in I_2 | H_{12}p_2 \in I_1 \right\}$.
The \emph{repeatability} rate $r$ is defined as:
\begin{equation}
\label{jk:eq:repeat}
r_{1,2} = \frac{|C(I_1, I_2)|}
{\min \left( \left| d_1 \right|, \left| d_2 \right| \right)} 
\quad ,
\end{equation}
where $C(I_1, I_2) = \left\{ 
\left( p_1 \in I_1, p_2 \in I_2 \right) 
|
dist(H p_1, p_2) < \theta
\right\}$ 
is the set of corresponding pairs of interest points.

In general case, when interest points are not related by homography, the above approach cannot be taken.
If the dataset ground truth contains image poses and high quality depth maps, 
as is the case in ApolloScape~\cite{huang2018apolloscape} dataset,
we can use it to establish the correspondence between interest points.
The 3D point $P$ corresponding to the interest point $p_1 \in I_1$ has the coordinates $P = (d \widetilde{x}, d \widetilde{y}, d)$, 
where $d$ is the depth of the point $p_1$
and $( \widetilde{x}$, $\widetilde{y} )$ are normalized coordinates of the point $p_1$ in the image $I_1$.
Then, 
using the ground truth relative pose between images $I_1$ and $I_2$,
we project $P$ onto the second image $I_2$ to get its projection $p^*_1$.
A pair of interest points $p_1 \in I_1$ and $p_2 \in I_2$ corresponds, if:
\begin{equation}
\label{jk:eq:corresponding_keypoints}
dist(p^*_1, p_2) < \theta. 
\end{equation}


\subsection{Evaluation results}

We evaluate stability of traditional interest point detectors: 
DoG~\cite{lowe1999object},
AKAZE~\cite{alcantarilla2011fast},
AGAST~\cite{mair2010adaptive},
Fast~\cite{rosten2006machine},
ORB~\cite{rublee2011orb}
and 
recently proposed detectors 
LIFT~\cite{yi2016lift},
Saddler~\cite{aldana2016saddle},
Superpoint~\cite{detone2017superpoint},
TILDE~\cite{verdie2015tilde}
and
LF-Net~\cite{ono2018lf}.

For traditional detectors (DoG, 
AKAZE,
AGAST,
Fast and ORB) we use OpenCV
\footnote{\url{https://opencv.org/}} implementations.
Evaluation of recent learning-based methods (LIFT, 
Superpoinit, TILDE, LF-Net) is performed using the code and pre-trained models released by authors.

\begin{table}[t]
\begin{center}

\caption{Mean repeatability of interest point detectors evaluated on 9 traversals from ApolloScape dataset.
Recently proposed Saddler detector has the highest average repeatability (0.177) for all evaluated sequences.
Traditional FAST detector is a runner up with 0.164 average repeatability.
The best deep-learning based detector Superpoint scores only 0.123.
}
\label{jk:repeatability_1seq}
\begin{tabular}{l c c c c c c c c c c}
\hline
Traversal id / &  1-16-6 & 1-32-5 & 1-36-6 & 2-01-5 & 2-05-5 & 2-16-5 & 3-01-5 & 3-21-5 & 3-29-5 & \textbf{Avg.} \\
Keypoint & \multicolumn{10}{c}{Repeatability} \\
\hline
Saddler~\cite{aldana2016saddle} & 0.103 & 0.089 & 0.153 & \textbf{0.186} & \textbf{0.090} & \textbf{0.231} & \textbf{0.247} & \textbf{0.255} & \textbf{0.238} & \textbf{0.177} \\
FAST~\cite{rosten2006machine} &	\textbf{0.113} & \textbf{0.093} & \textbf{0.200} & 0.176 & 0.077 & 0.217 & 0.206 & 0.185 & 0.208 & 0.164 \\
AGAST~\cite{mair2010adaptive} & 0.091 & 0.086 & 0.176 & 0.146 & 0.070 & 0.190 & 0.179 &	0.163 & 0.182 & 0.143 \\
ORB~\cite{rublee2011orb}	& 0.078 & 0.056 & 0.138 & 0.140 & 0.038 & 0.192 & 0.236 & 0.196 & 0.207 & 0.142 \\
AKAZE~\cite{alcantarilla2011fast} & 0.081 & 0.070 & 0.169 & 0.142 & 0.072& 0.176 & 0.184 & 0.160 & 0.177 & 0.137 \\
Superpoint~\cite{detone2017superpoint} & 0.075 & 0.070 & 0.119 & 0.136 & 0.080 & 0.152 & 0.168 & 0.156 & 0.151 & 0.123 \\
DoG~\cite{lowe1999object} & 0.081 & 0.066 & 0.122 & 0.137 & 0.062 & 0.156 & 0.161 & 0.143 & 0.168 & 0.122 \\
TILDE~\cite{verdie2015tilde} & 0.068 & 0.068 & 0.088 & 0.092 & 0.074 & 0.107 & 0.108 & 0.103 & 0.116 & 0.091 \\
LF-Net~\cite{ono2018lf} & 0.060 & 0.065 & 0.095 & 0.082 & 0.034 & 0.104 & 0.114 & 0.101 & 0.103 & 0.084 \\
LIFT~\cite{yi2016lift} & 0.050 & 0.052 & 0.054 & 0.058 & 0.049 & 0.056 & 0.068 & 0.061 & 0.072 & 0.058 \\
\hline
\end{tabular}
\end{center}
\end{table}


We choose the following traversals
from ApolloScape~\cite{huang2018apolloscape} dataset: 
16, 32 from road id 1;
1, 5, 16 from road id 2;
1, 21 and 29 from road id 3.
For each traversal we take every 20th frame (corresponding to 20 meter drive) as the base frame.
We match keypoints in each base frame with keypoints in 19 subsequent frames. These 19 subsequent frames are at the distance of approximately $1, 2, \ldots, 19$ meters from the base frame.
On each image we select $N=10,000$ interest points with the strongest response.
The threshold $\theta$ for interest point correspondence in Eq.~\ref{jk:eq:corresponding_keypoints} is set to 2.5 pixels.

\begin{figure}
\centering
\includegraphics[width=0.49\textwidth]{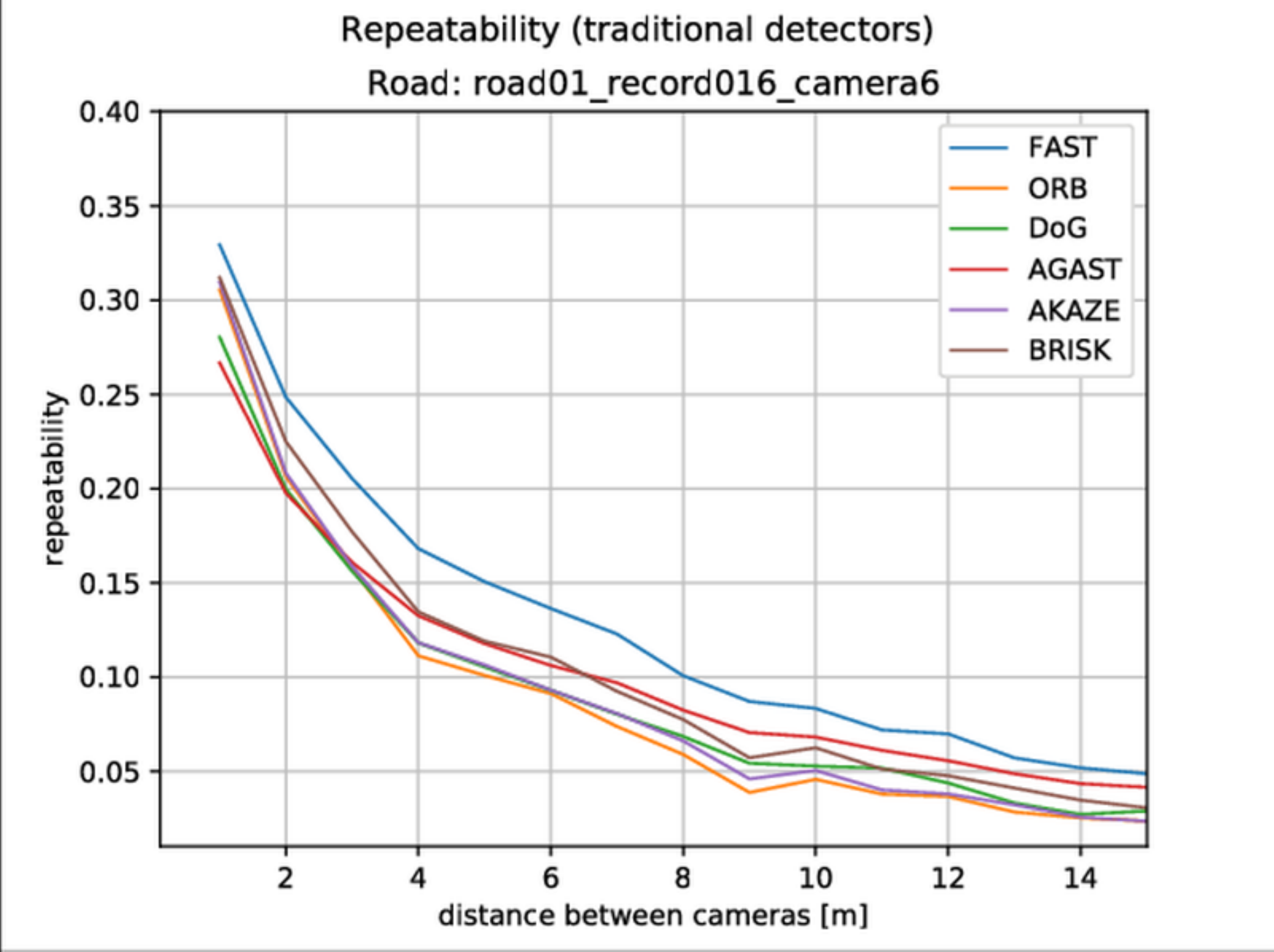}
\includegraphics[width=0.49\textwidth]{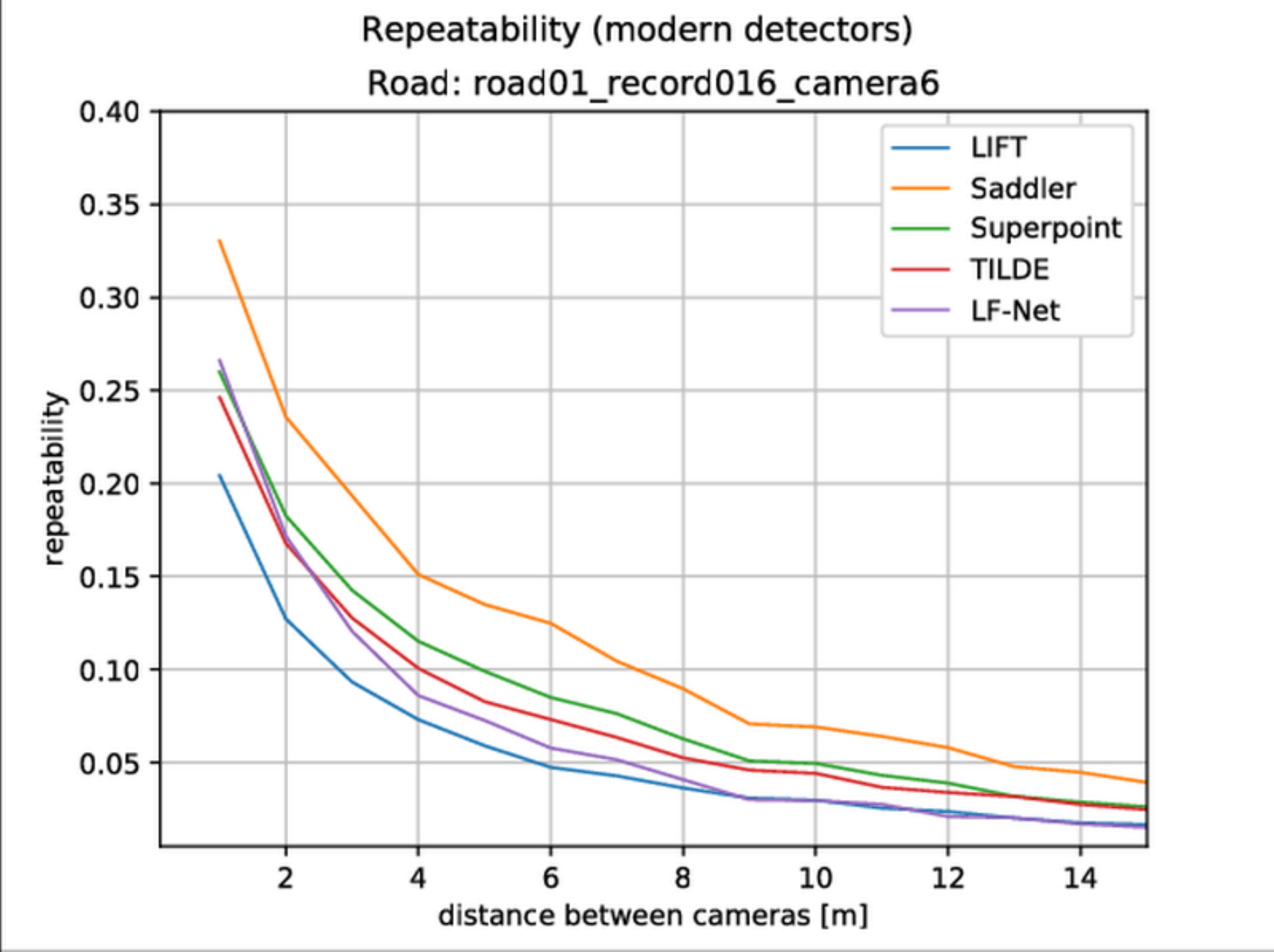}
\includegraphics[width=0.49\textwidth]{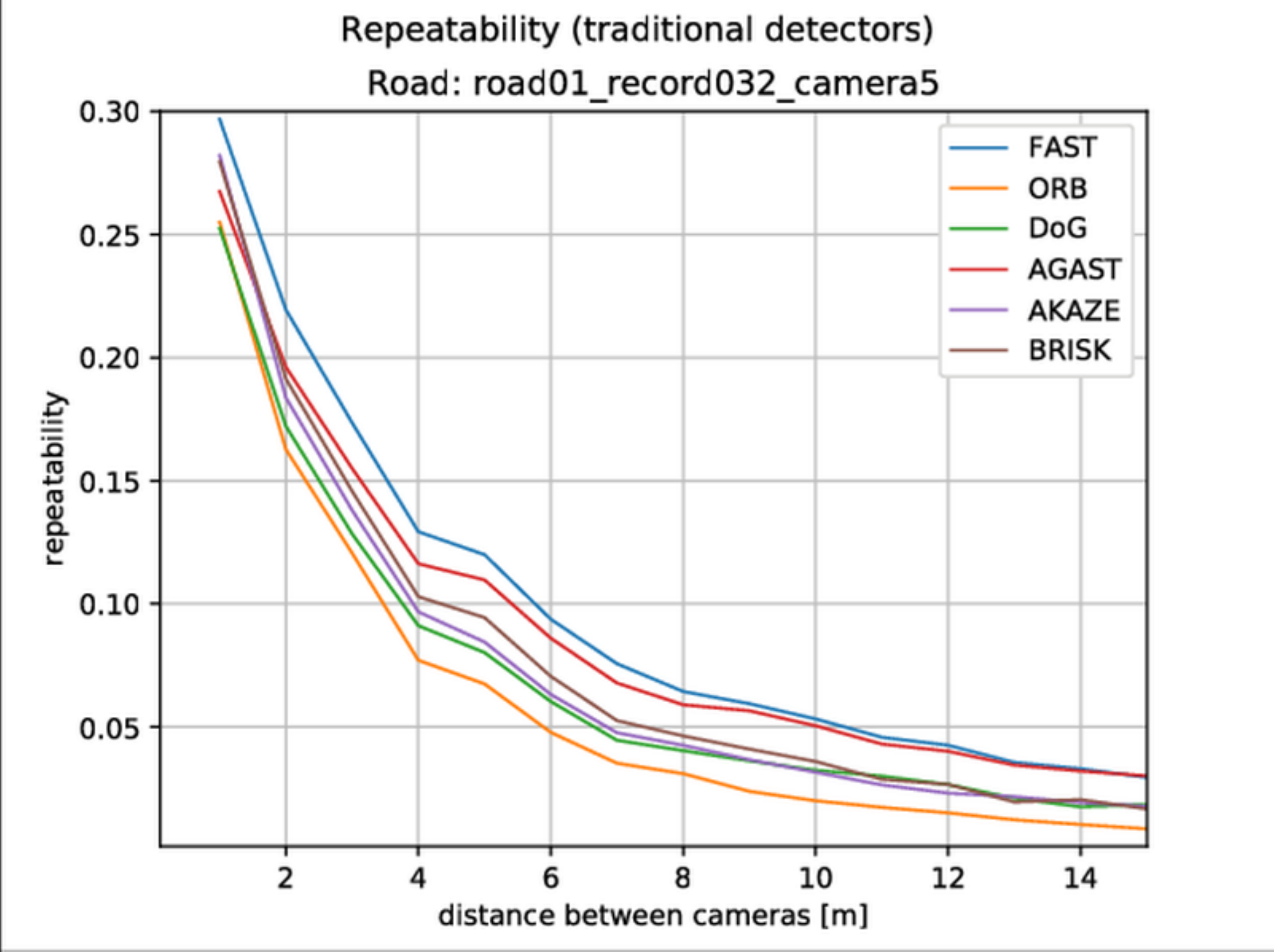}
\includegraphics[width=0.49\textwidth]{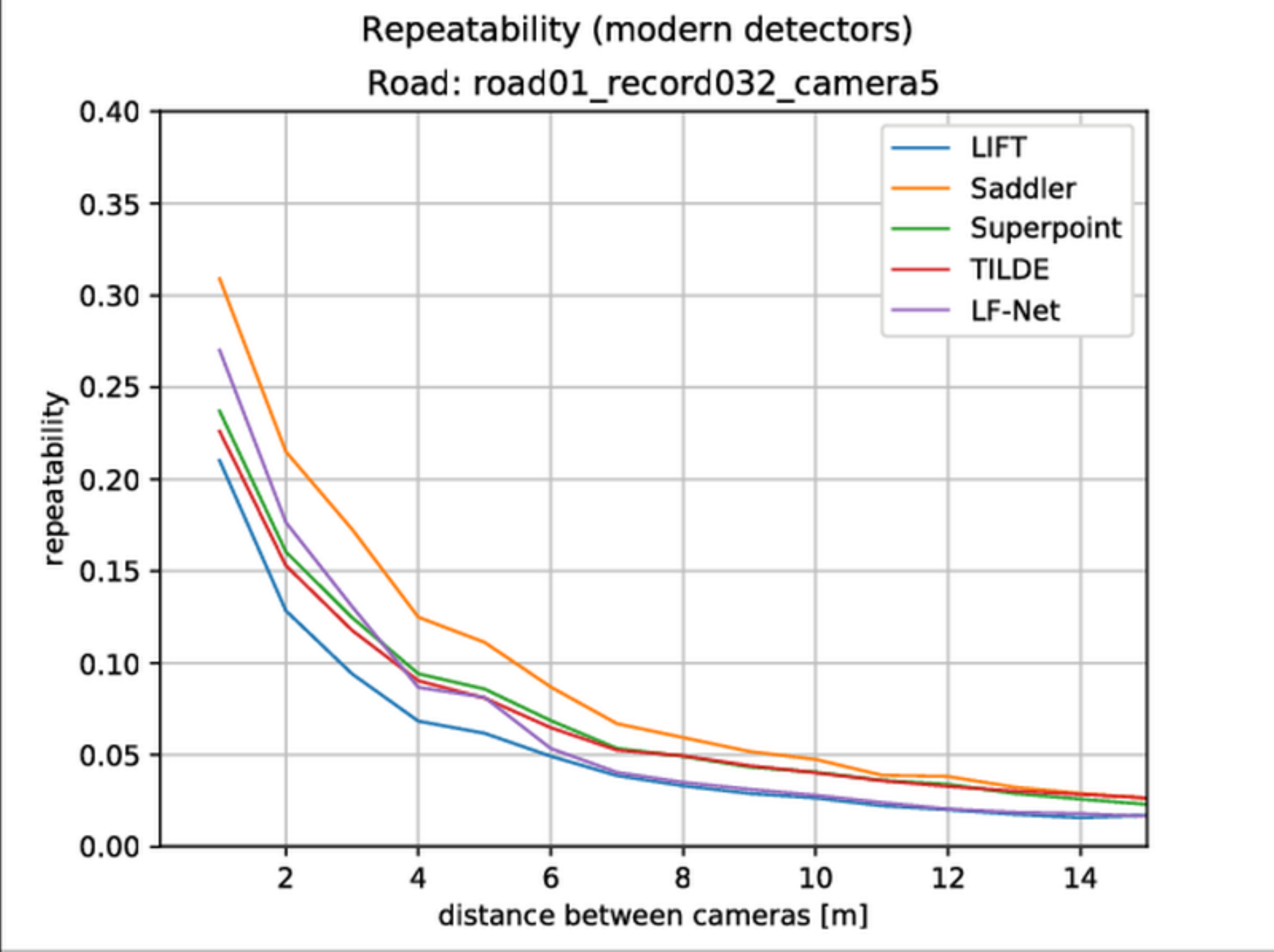}
\includegraphics[width=0.49\textwidth]{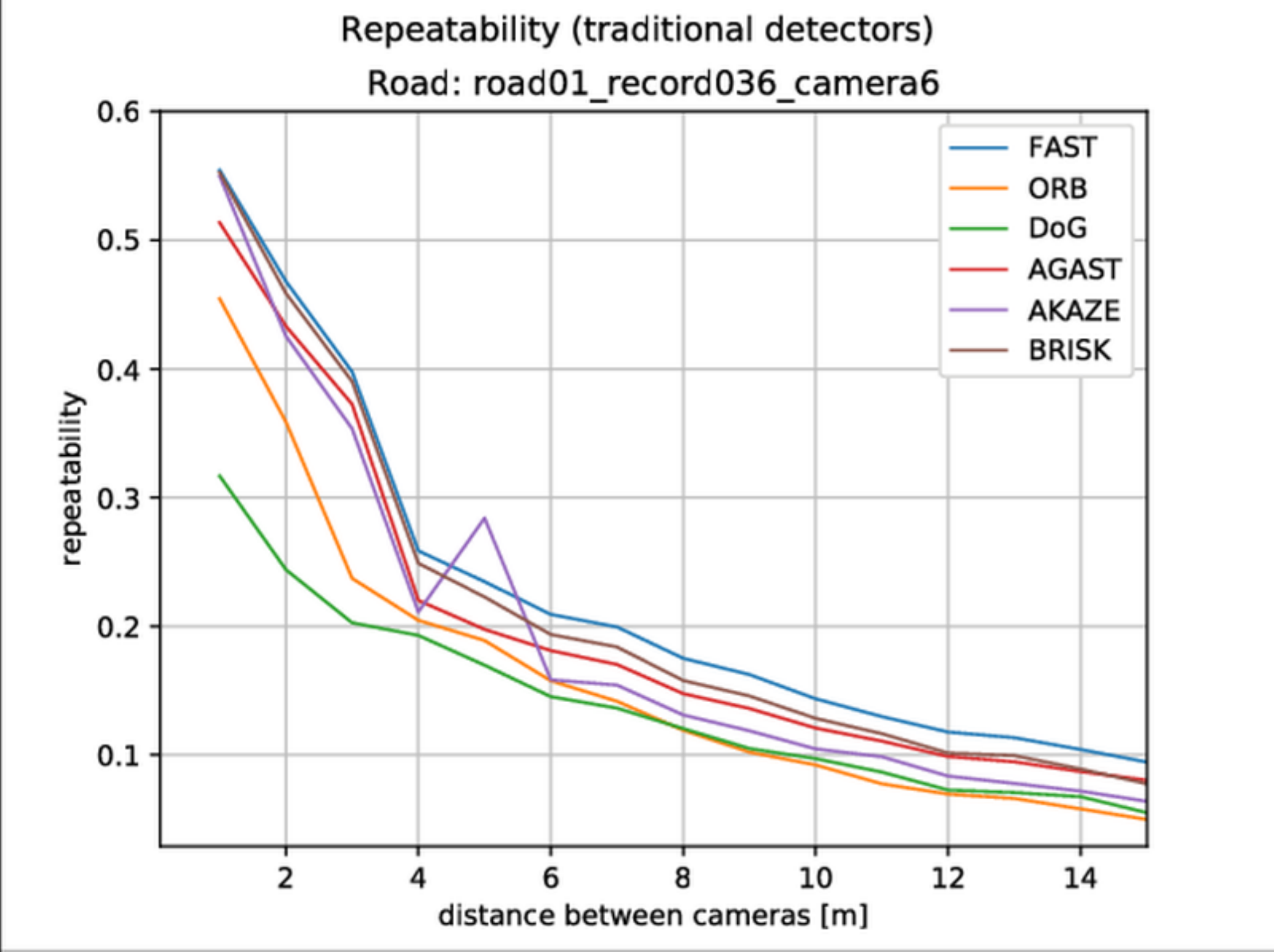}
\includegraphics[width=0.49\textwidth]{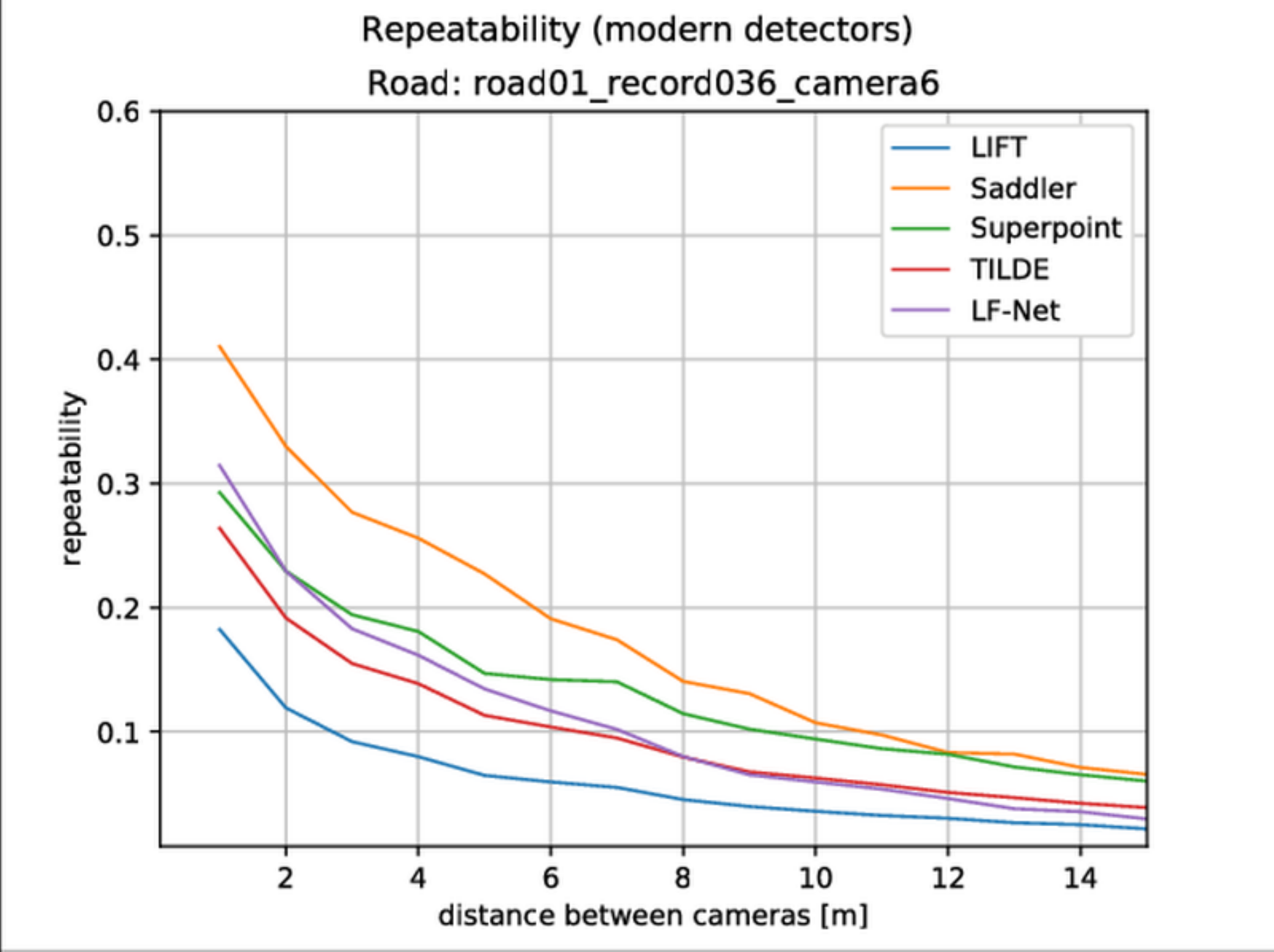}
\caption{Repeatability of keypoint detectors evaluated on record 016 (top), record032 (middle) and record 036 (bottom) from road 01
as the function of distance between cameras. 
(Left column) Traditional keypoint detectors,
(right column) modern keypoint detectors.
}
\label{fig:repeatability_apollo1}
\end{figure}

\begin{figure}
\centering
\includegraphics[width=0.49\textwidth]{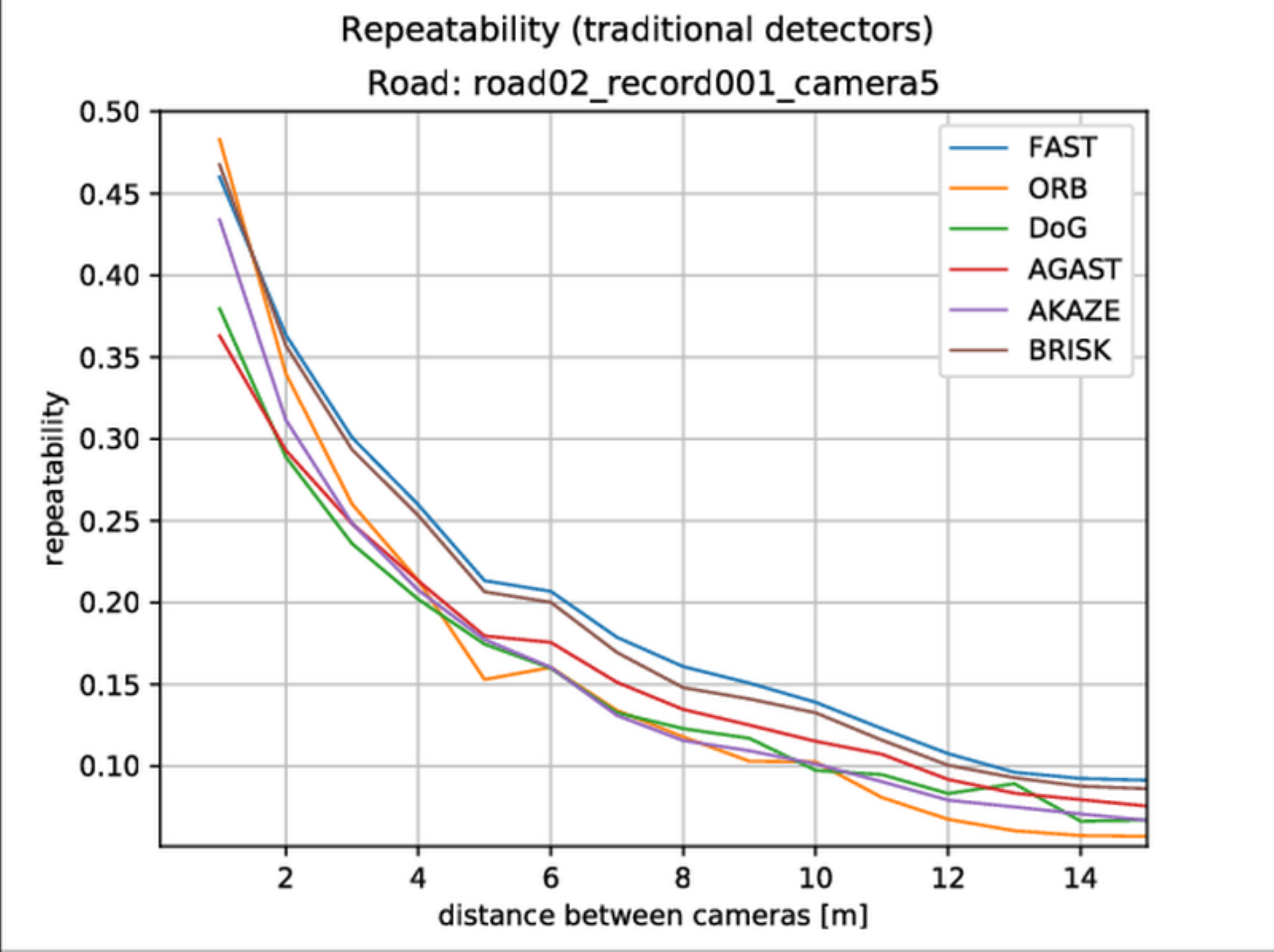}
\includegraphics[width=0.49\textwidth]{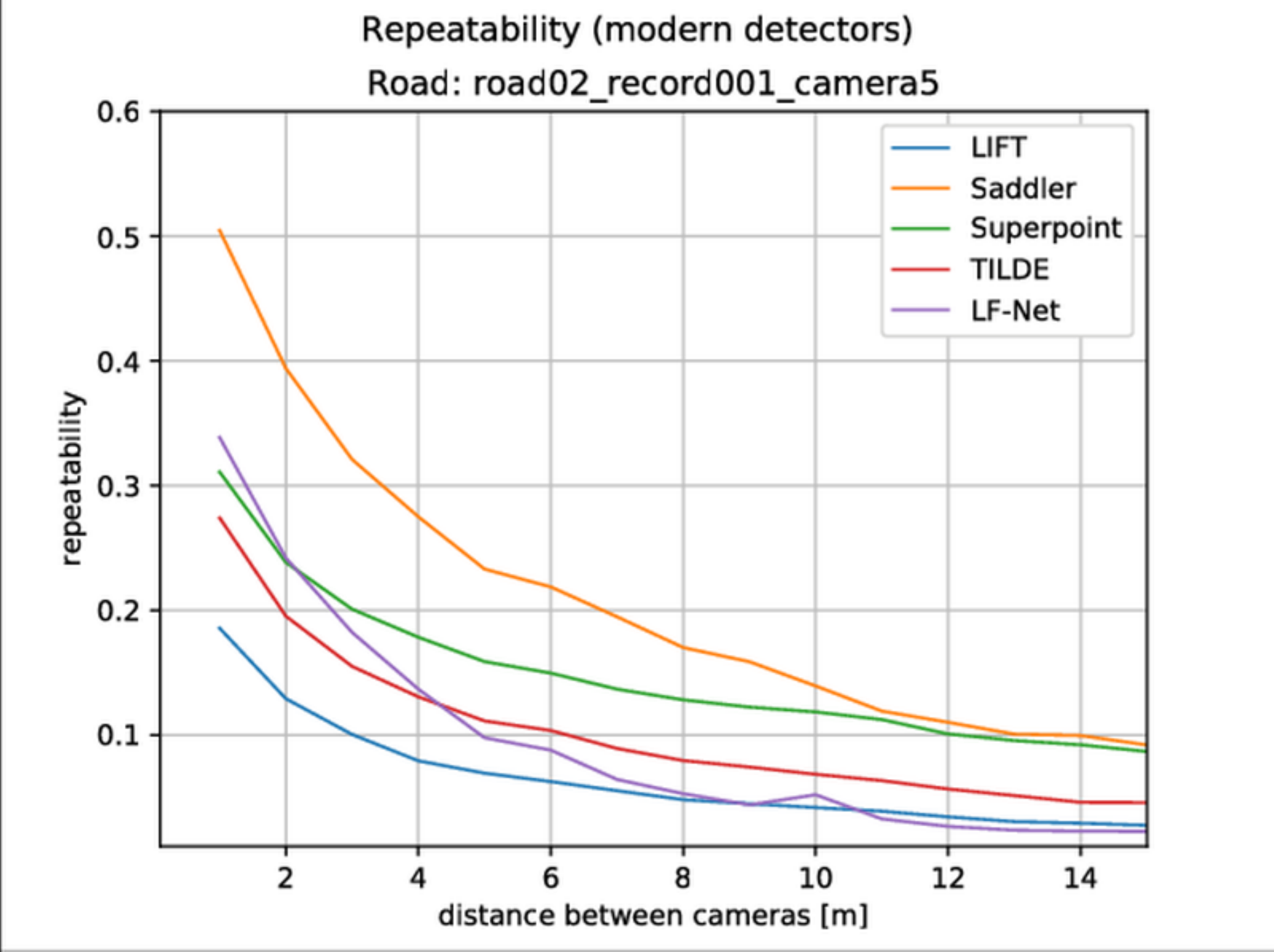}
\includegraphics[width=0.49\textwidth]{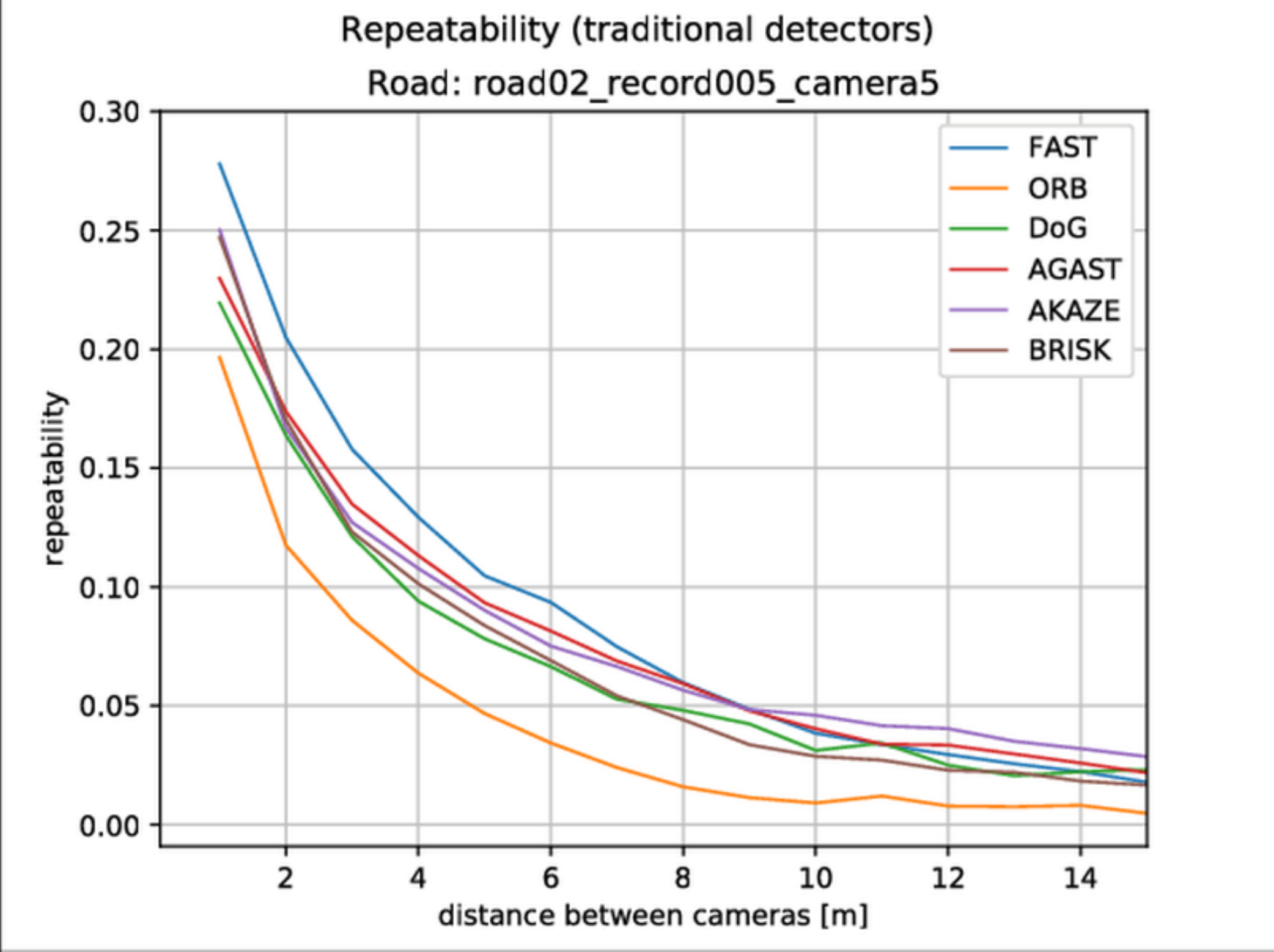}
\includegraphics[width=0.49\textwidth]{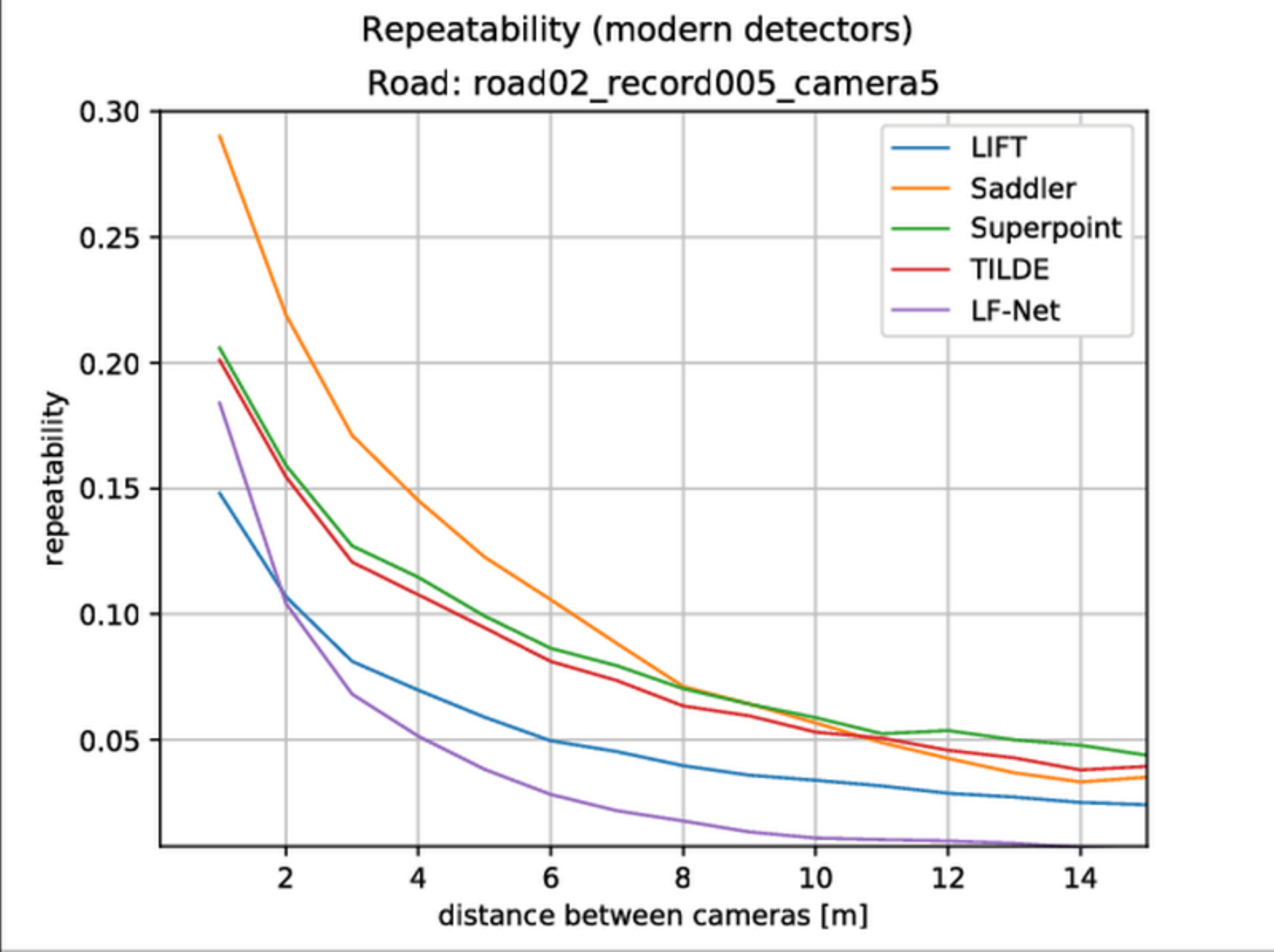}
\includegraphics[width=0.49\textwidth]{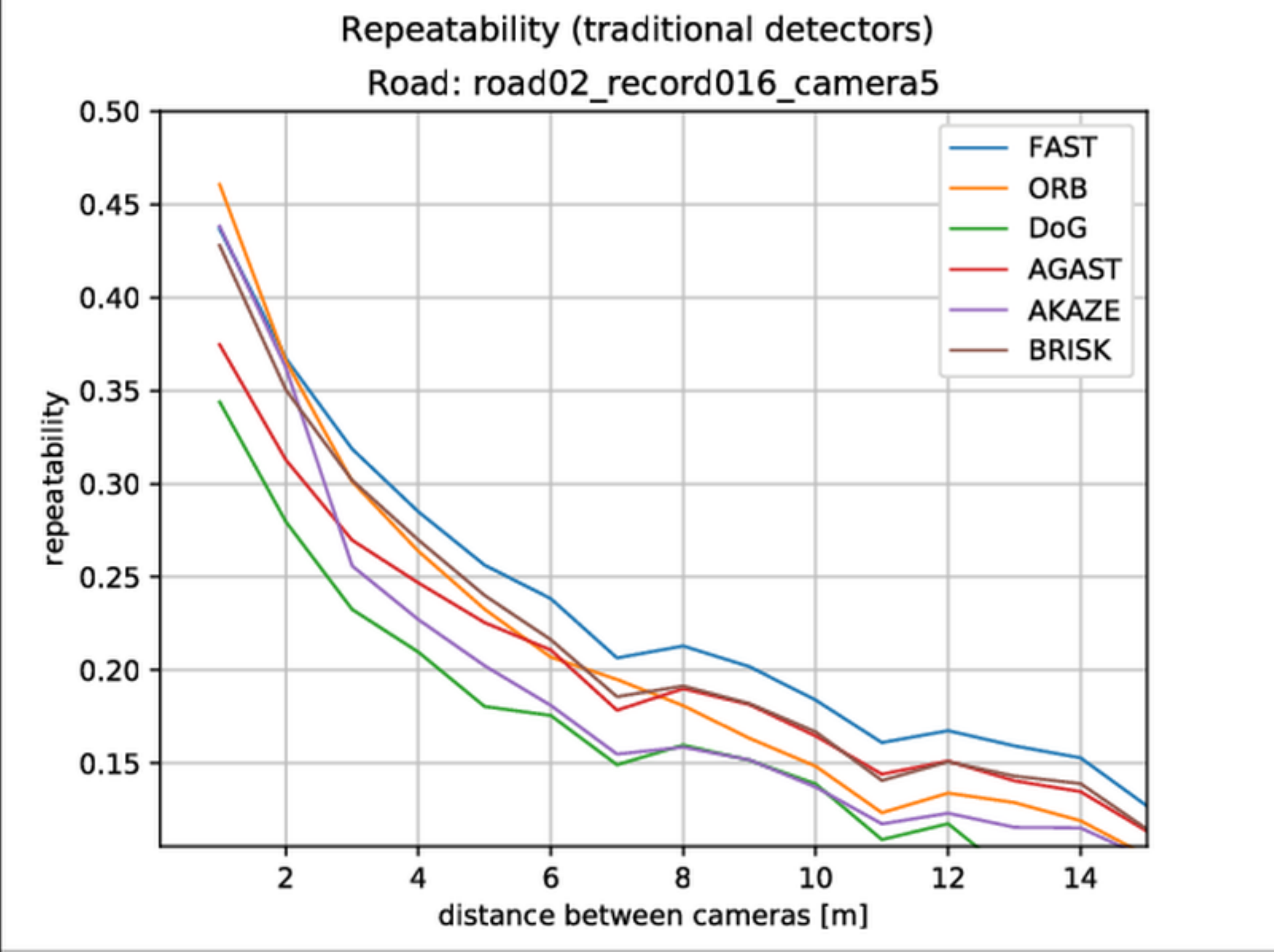}
\includegraphics[width=0.49\textwidth]{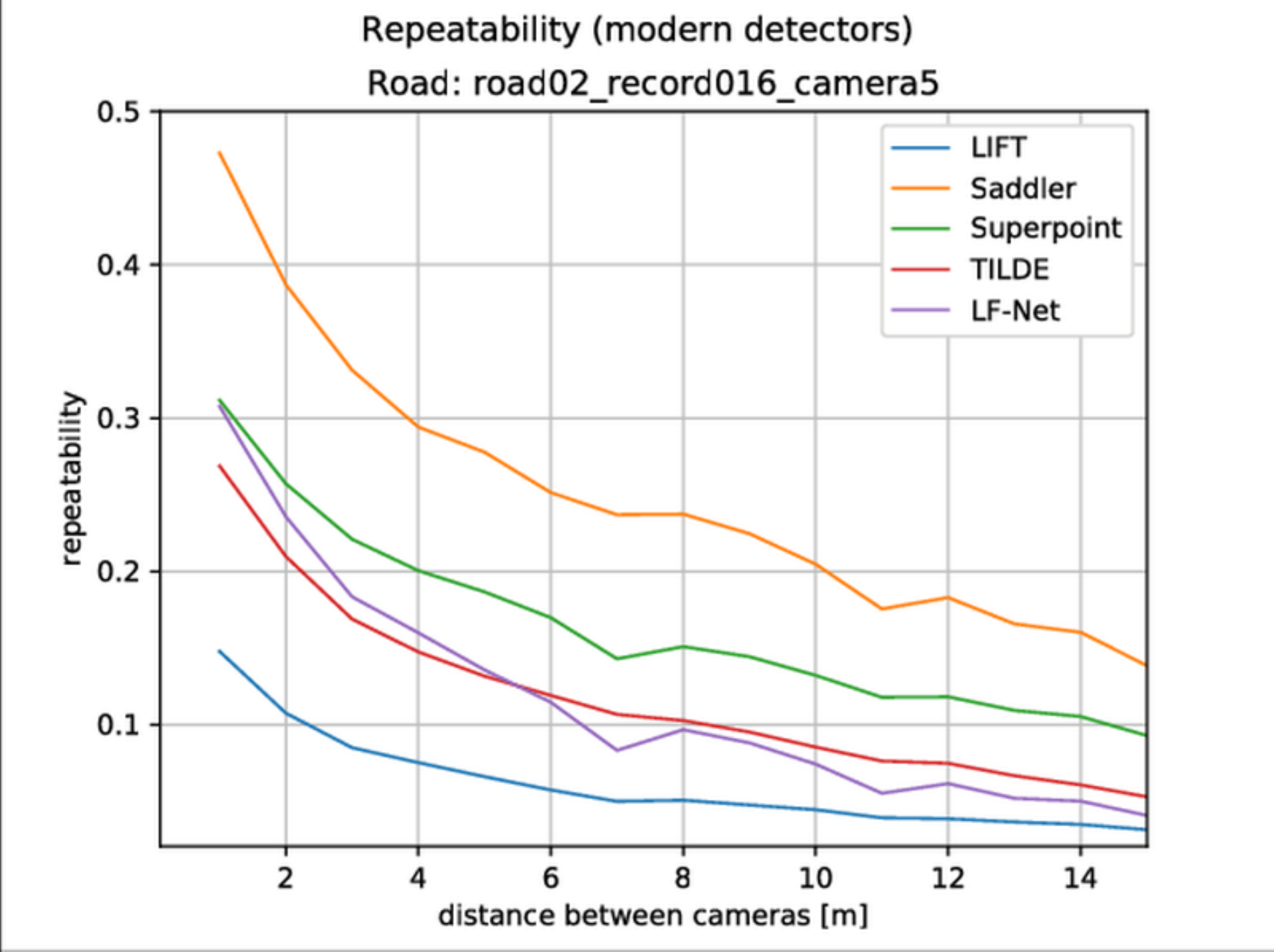}
\caption{Repeatability of keypoint detectors evaluated on record 001 (top), record005 (middle) and record 016 (bottom) from road 02
as the function of distance between cameras. 
(Left column) Traditional keypoint detectors,
(right column) modern keypoint detectors.
}
\label{fig:repeatability_apollo2}
\end{figure}

\begin{figure}
\centering
\includegraphics[width=0.49\textwidth]{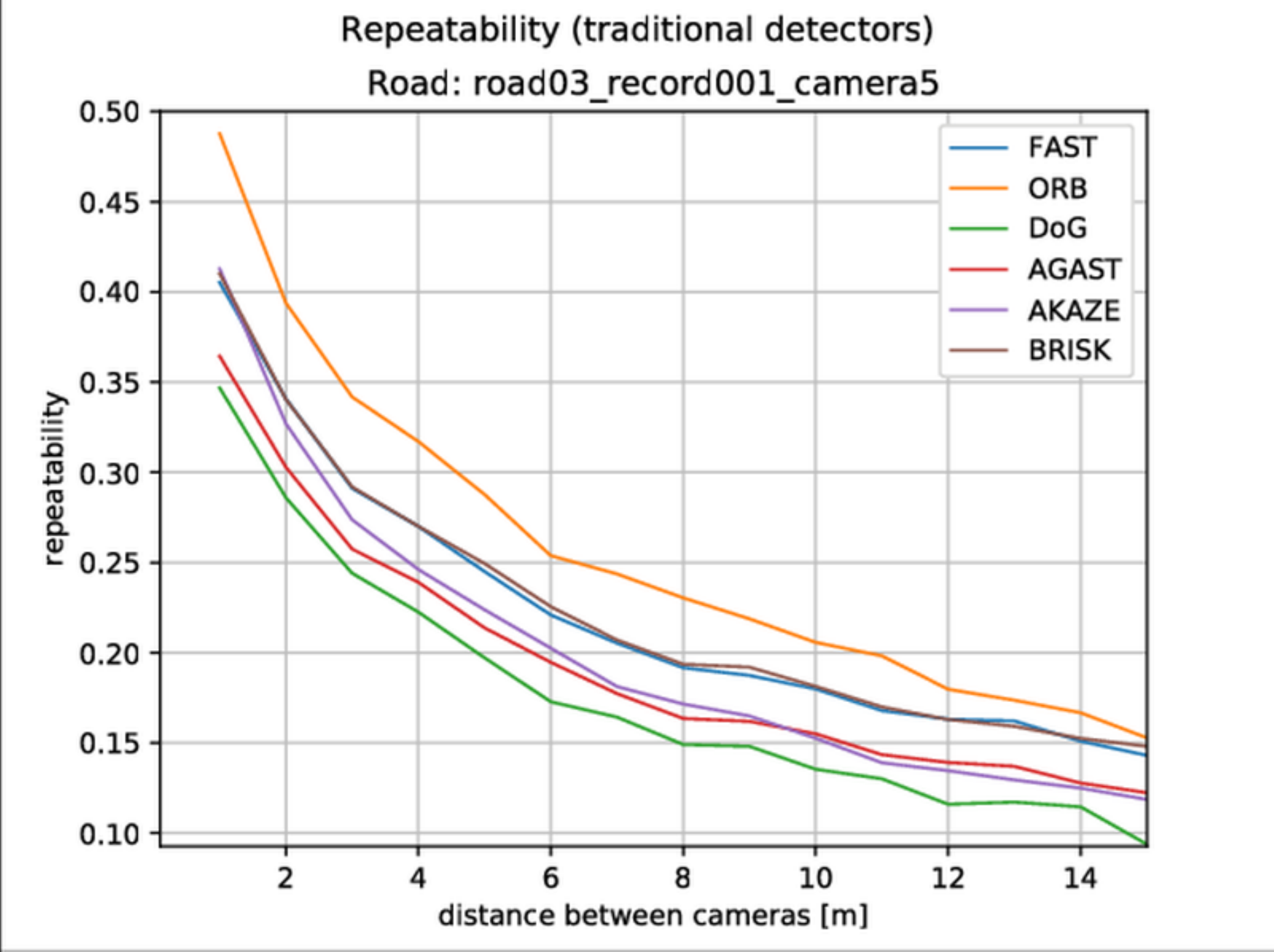}
\includegraphics[width=0.49\textwidth]{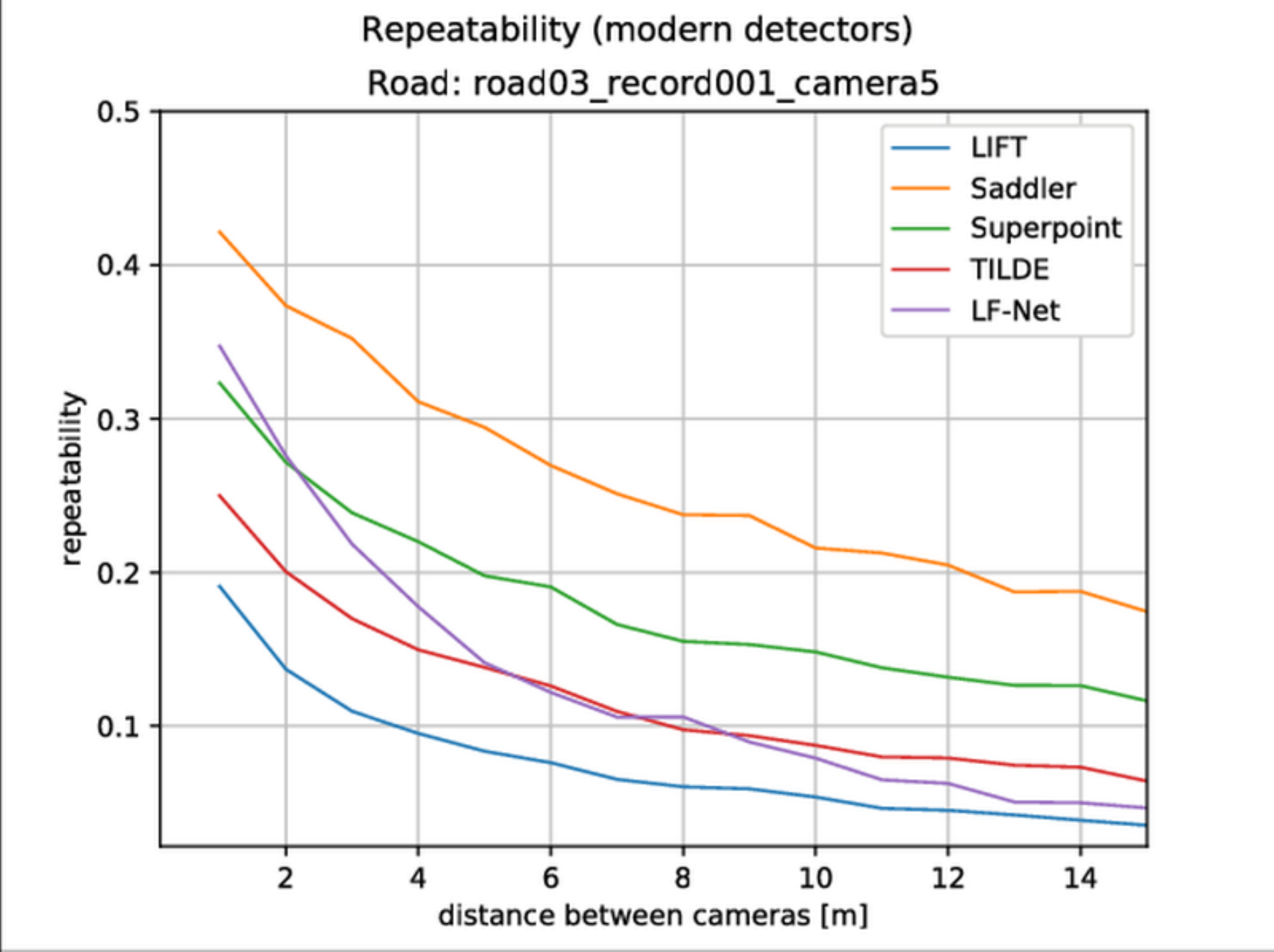}
\includegraphics[width=0.49\textwidth]{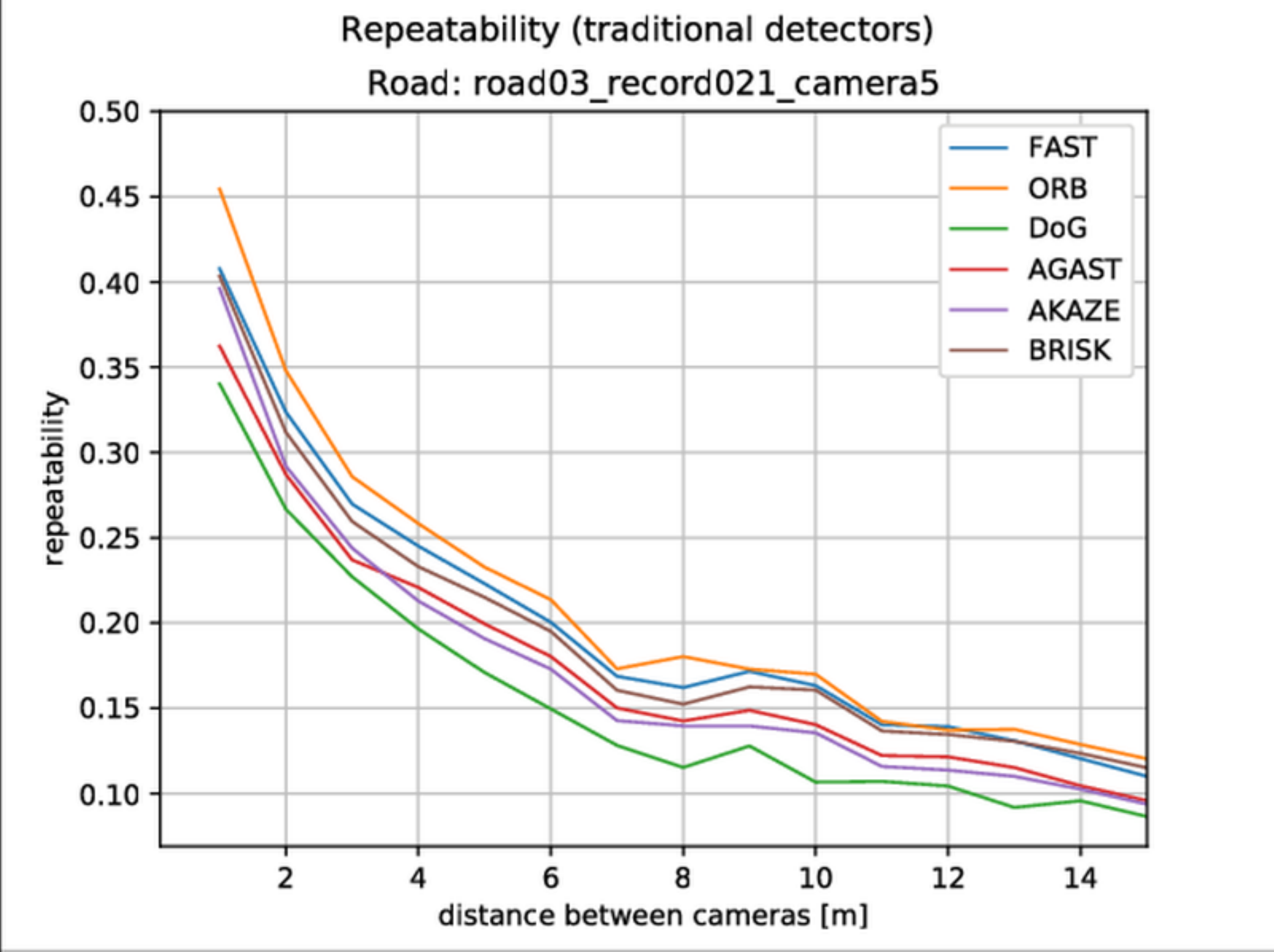}
\includegraphics[width=0.49\textwidth]{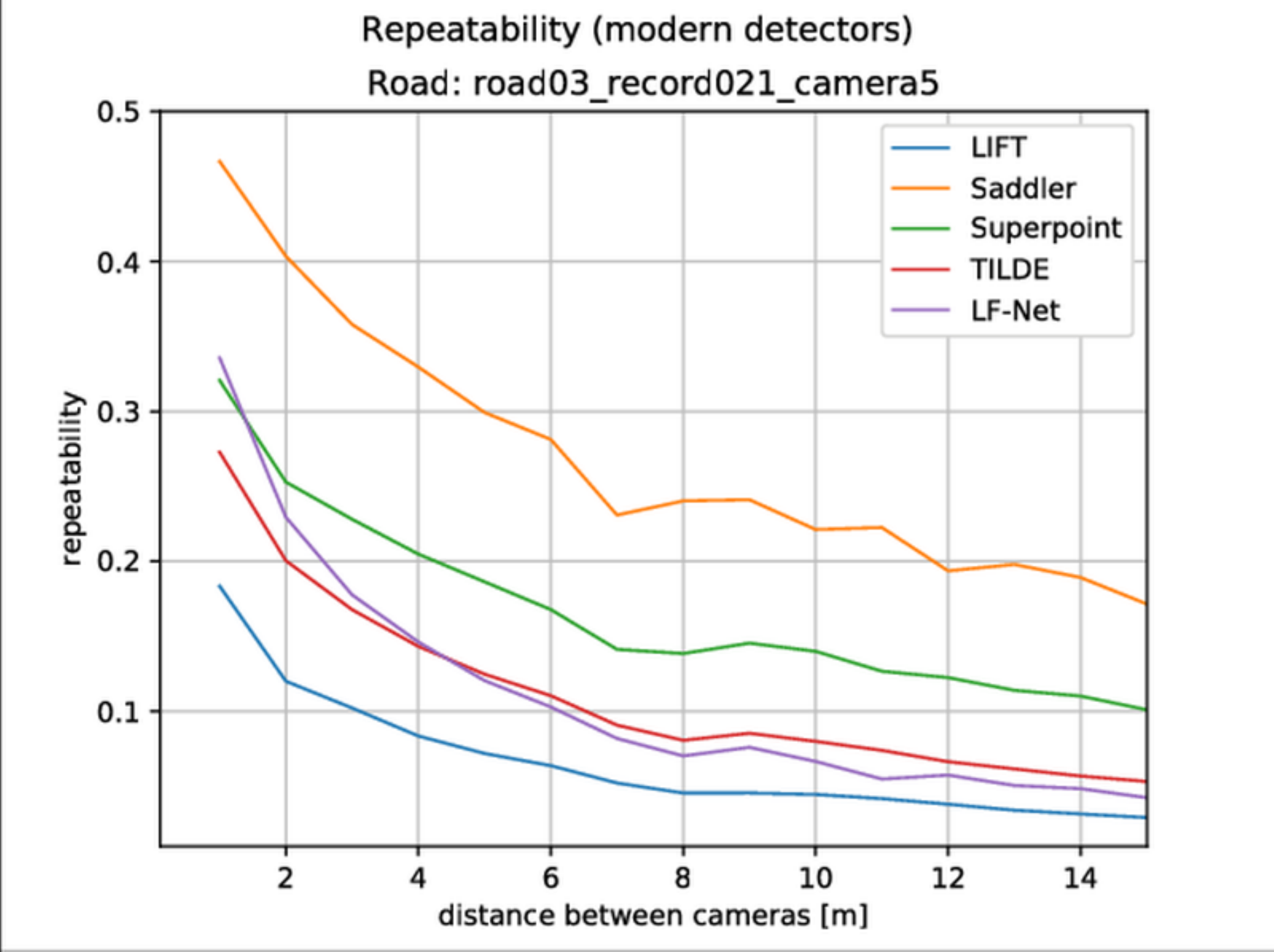}
\includegraphics[width=0.49\textwidth]{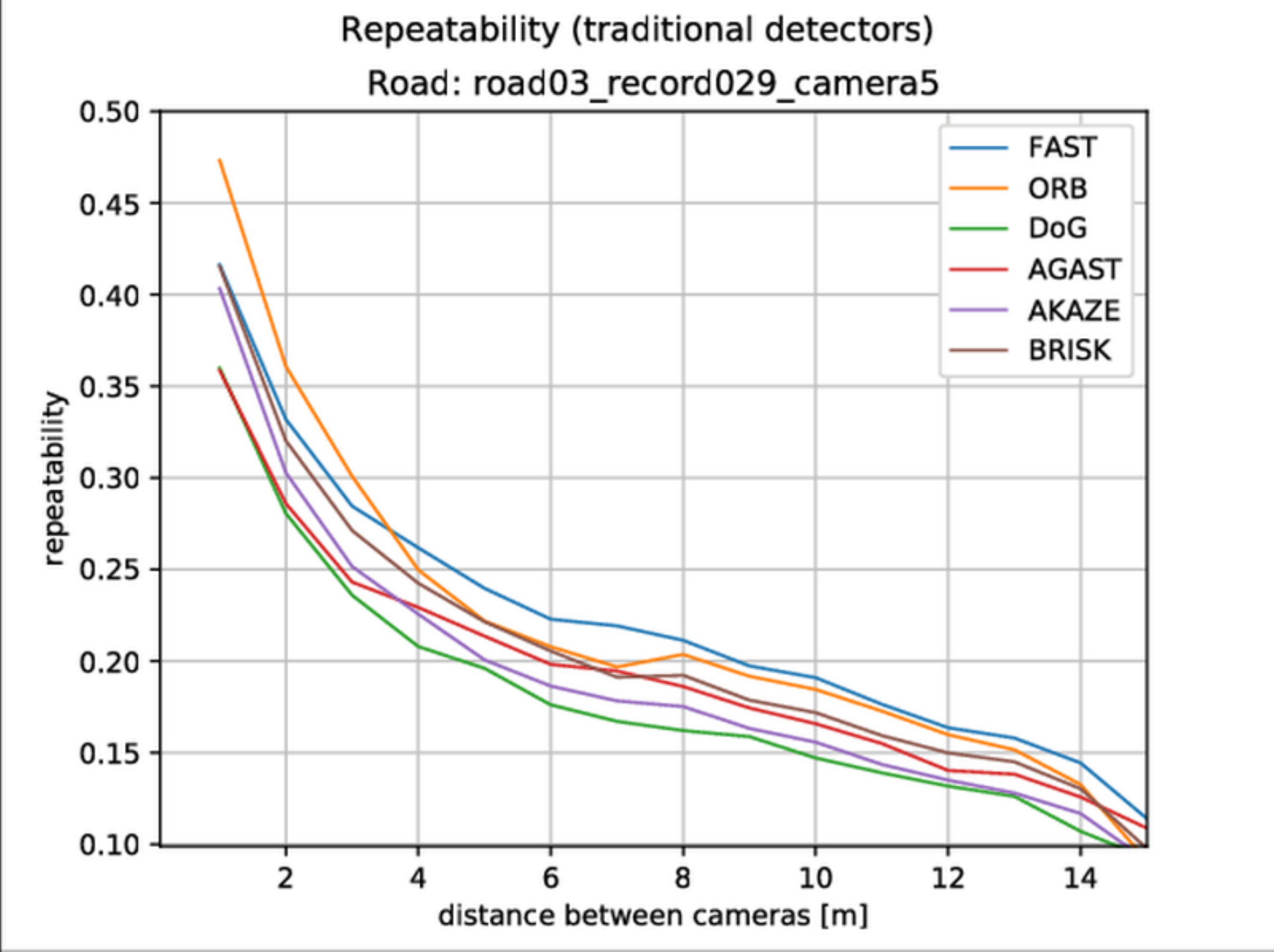}
\includegraphics[width=0.49\textwidth]{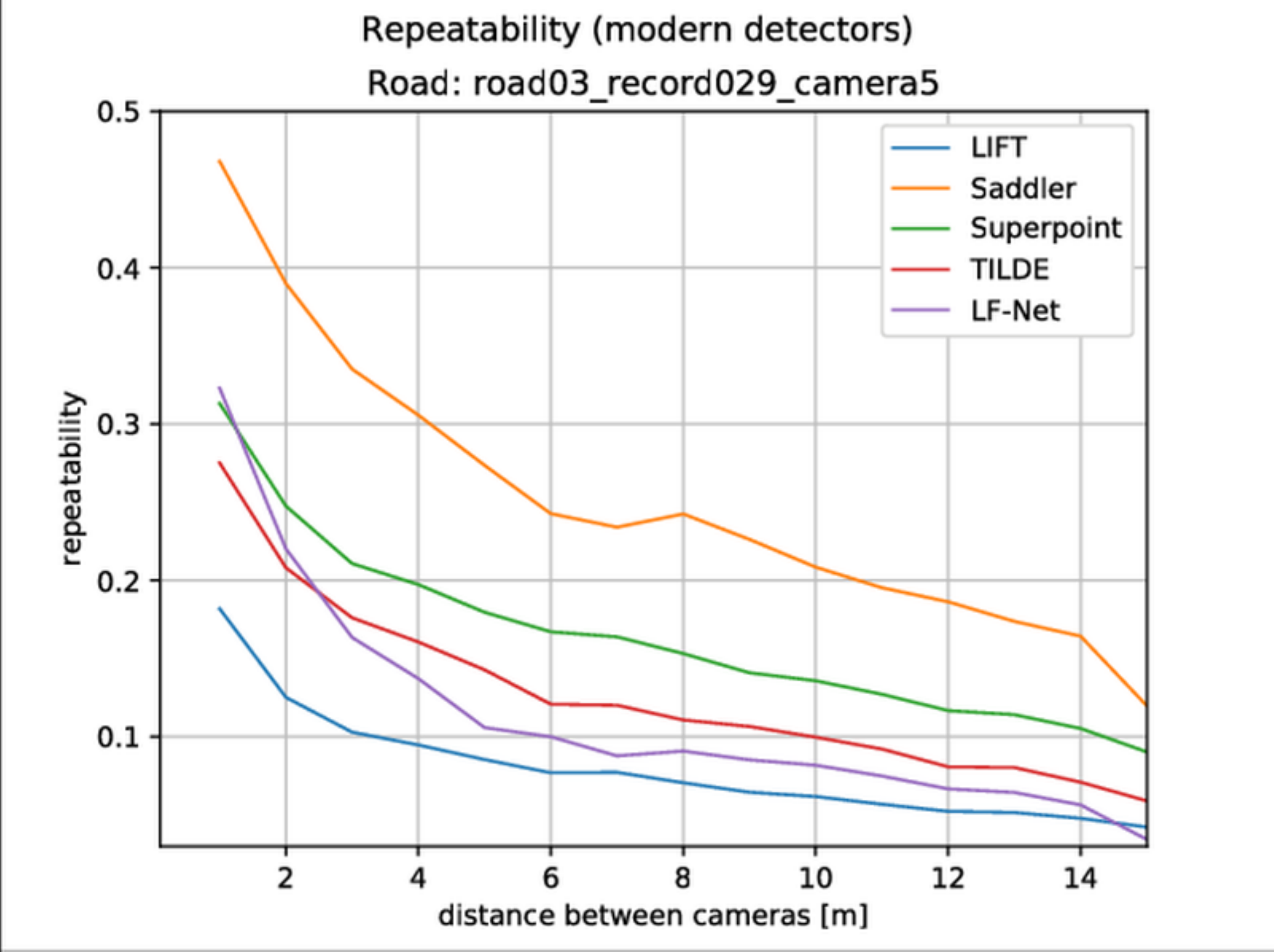}
\caption{Repeatability of keypoint detectors evaluated on record 001 (top), record021 (middle) and record 029 (bottom) from road 3
as the function of distance between cameras. 
(Left column) Traditional keypoint detectors.
(right column) modern keypoint detectors.
}
\label{fig:repeatability_apollo3}
\end{figure}

Fig.~\ref{fig:repeatability_apollo1}, \ref{fig:repeatability_apollo2} and \ref{fig:repeatability_apollo3}
show mean repeatability score as a function of a distance between camera centres in pairs of images for each evaluated traversal.
The results are summarized in Tab.~\ref{jk:repeatability_1seq}.

Recently proposed Saddler~\cite{aldana2016saddle} detector has the best average repeatability for all evaluated traversals: 0.177. 
It has the highest repeatability in 6 out of 9 traversals and a relatively small gap to the best performing detectors in the remaining 3 traversals.
Traditional FAST~\cite{rosten2006machine} (avg. repeatability 0.164) performs slightly worse.

Non of the recently proposed deep learning-based keypoint detectors shows an advantage other the traditional hand-crafted detectors.
The best performing deep-learning based detector, Superpoint~\cite{detone2017superpoint}, has an average repeatability equal to 0.123.
Other perform even worse: TILDE~\cite{verdie2015tilde} scores 0.084, LF-Net~\cite{ono2018lf} 0.04 and LIFT~\cite{yi2016lift} 0.058 average repeatability.

\begin{figure}
\centering
\includegraphics[width=0.8\textwidth]{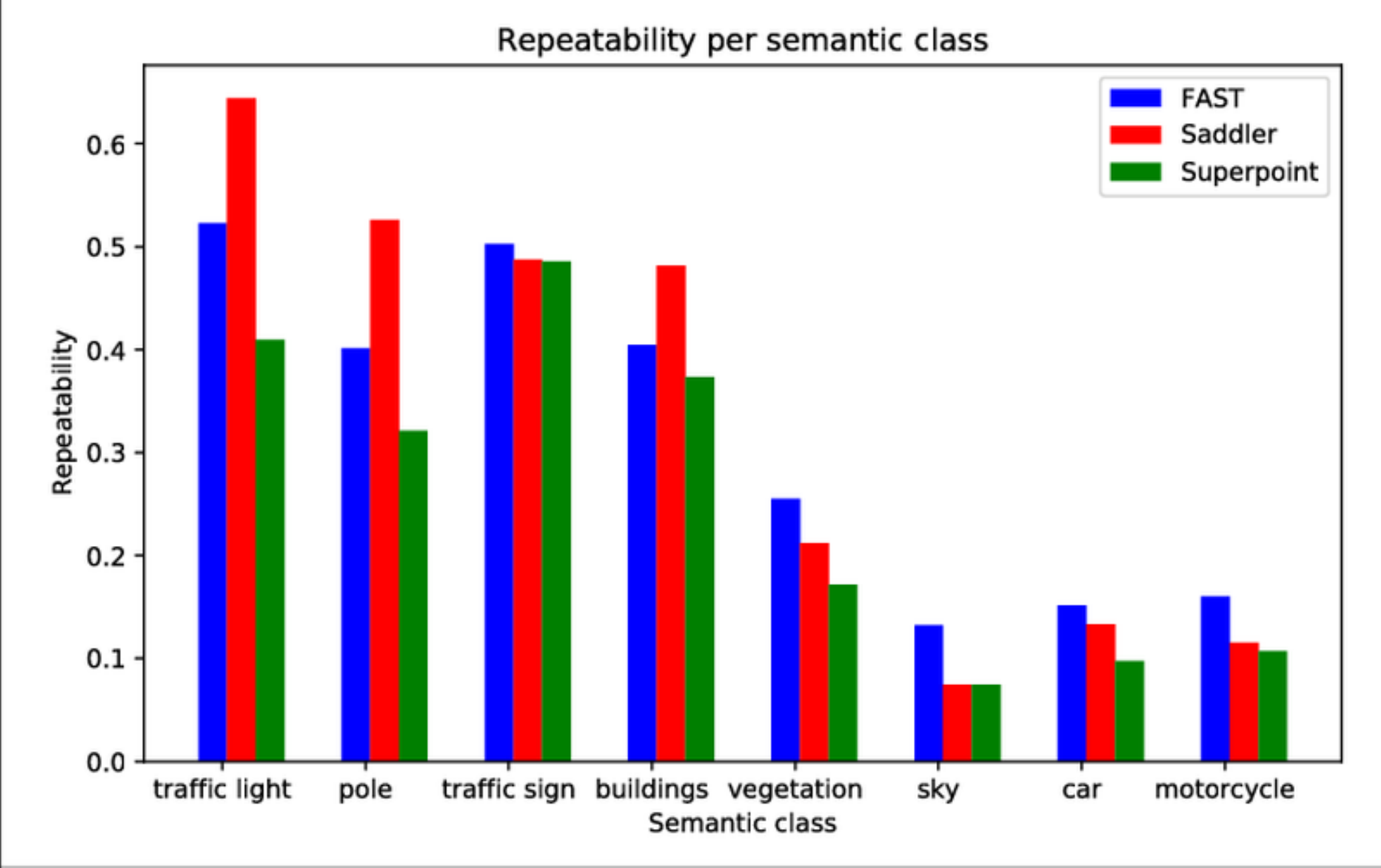}
\caption{Repeatability of FAST, Saddler and Superpoint keypoints on objects with different semantic labels. 
Repeatability for stable objects such as traffic signs or buildings (left side) is consistently the highest for all detectors.
Repeatability for movable objects (right side) is the lowest.
}
\label{fig:repeatability_semantic}
\end{figure}

Using semantic labels ground truth provided with the ApolloScape dataset, we analyzed dependency of the interest point repeatability on the semantic content of the visible scene.
Tab.~\ref{jk:repeatability_perclass} shows semantic classes with top-5 and bottom-5 repeatability for the best performing detector (Saddler~\cite{aldana2016saddle}), best traditional detector (FAST~\cite{rosten2006machine}) and best deep-learning based detector (Superpoint~\cite{detone2017superpoint}).
As expected, repeatability of keypoints detected on stable objects such as billboards, traffic lights and buildings is consistently higher than on movable or volatile objects such as car, motorcycles, or sky.
Dependency of the interest point detector performance on the semantic class of the scene region is visualized on Fig.~\ref{fig:repeatability_semantic}.
These results proves, that in order to build high-performing interest point detector, high-level information on the semantic content of the visible scene must be taken into the account.
Keypoints should not be detected on the movable or volatile regions of the scene, such as vehicles or sky. 

\begin{table}[t]
\begin{center}

\caption{
Semantic classes with top-5 and bottom-5 repeatability for FAST, Saddler and Superpoint keypoint detectors.
Repeatability of keypoints detected on the stable structures such as billboards, traffic light or traffic signs) is significantly higher than repeatability of the keypoints detected on movable or volatile objects, such as cars, motorcycles or sky.
}
\label{jk:repeatability_perclass}
\begin{tabular}{l l c p{10pt} l c}
\hline
 & \multicolumn{2}{c}{Top 5} & & \multicolumn{2}{c}{Bottom 5} \\
Detector  &  Semantic class & Avg. rep. & & Semantic class & Avg. rep. \\
\hline
FAST~\cite{rosten2006machine} & billboard & 0.552 & & rover  & 0.021\\
& traffic light  & 0.523 & & sky  & 0.132 \\
& traffic sign  & 0.502 & & unlabelled  & 0.138 \\
& road pile  & 0.491	&  & car  & 0.151 \\
& dustbin  & 0.418 & & motorcycle &  0.160 \\
\\
Saddler~\cite{aldana2016saddle} & traffic light & 0.640 & & rover & 0.012 \\
& billboard & 0.538 & & sky  & 0.075 \\
& pole  & 0.526 & & motorcycle  & 0.115 \\
& traffic sign  & 0.487 & & person  & 0.130 \\
& building  & 0.482 & &  car  & 0.133 \\
\\
 Superpoint~\cite{detone2017superpoint} & billboard  & 0.538 & & rover  & 0.030 \\
& traffic sign  & 0.486 & & sky  & 0.075 \\
& road pile   &	0.395 & & unlabelled  & 0.086 \\
& overpass  & 0.381 & & car  & 0.097 \\
& building  & 0.373 & &  motorcycle  & 0.107 \\
\hline
\end{tabular}
\end{center}
\end{table}

\section{Conclusions and future work}
\label{sec:conclusions}

The best performing interest point detector on ApolloScape dataset is recently proposed 
Saddler~\cite{aldana2016saddle}). 
Average Saddler repeatability across all evaluated sequences is 0.177. It has a considerable advantage over the second best detector, FAST (0.164 mean repeatability).

Interestingly, all recently proposed deep learning-based interest point detectors perform noticeable worse compared to best hand-crafted detector (Saddler).
The best deep learning-based detector, Superpoint~\cite{detone2017superpoint},  scores average  repeatability equal to 0.123. 
Other evaluated deep learning-based descriptors (TILDE~\cite{verdie2015tilde}, LF-Net~\cite{ono2018lf}, LIFT~\cite{yi2016lift}) perform even worse.

Analysis of the stability of interest points detected in objects with different semantic class labels proves that keypoints detected on the stable objects, such as road signs or buildings, have the highest repeatability.
This suggest, that in order to build high-performing interest point detector, high-level information on the semantic content of the visible scene must be taken into the account.
Keypoints should not be detected on the movable or volatile objects, such as vehicles or sky. 

As the future work we plan to verify if the results generalize to other large-scale, stree-level view datasets, such as Oxford RobotCar~\cite{RobotCarDatasetIJRR} and Berkeley DeepDrive~\cite{yu2018bdd100k}.

A very promising direction of the research is development of the repeatable and discriminant learning-based interest point detector and descriptor.
Recently released large-scale street-view datasets contain millions of images taken in multiple locations in diverse weather conditions, and at the different times of the day and year. 
Ground truth, in the form of high quality dense depth maps or point clouds, allows establishing correspondences between interest point in different images,
This should allow training high quality interest point detectors and descriptors, invariant to viewpoint change and variable environmental conditions.


\section*{Acknowledgement}
This research was supported by Google Sponsor Research Agreement under
the project "Efficient visual localization on mobile devices".

The Titan X Pascal used for this research was donated by the NVIDIA Corporation.

\clearpage

\bibliographystyle{splncs}
\bibliography{foxtrot-bib}
\end{document}